\documentclass{article}

\usepackage{amsmath}
\usepackage{amssymb}
\usepackage{algorithm}
\usepackage{algpseudocode}
\usepackage{hyperref}
\usepackage{graphicx} 
\usepackage{hyperref}

\usepackage[T1]{fontenc}
\usepackage[utf8]{inputenc}
\usepackage{authblk}

\author[1]{Kai-Chun Hu \thanks{hukaichun.am05g@nctu.edu.tw}}
\author[2]{Chen-Huan Pi\thanks{john40532.me00@g2.nctu.edu.tw}}
\author[3]{Ting Han Wei \thanks{tinghan.wei@gmail.com}}
\author[3]{I-Chen Wu\thanks{icwu@csie.nctu.edu.tw}}
\author[2]{Stone Cheng\thanks{stonecheng@mail.nctu.edu.tw}}
\author[2]{Yi-Wei Dai\thanks{be84n12son25@gmail.com}}
\author[2]{Wei-Yuan Ye\thanks{s10030789@gmail.com}}
\affil[1]{Department of Applied Mathematics, National Chiao Tung University}
\affil[2]{Department of Mechanical Engineering, National Chiao Tung University}
\affil[3]{Department of Computer Science, National Chiao Tung University}

\title{Towards Combining On-Off-Policy Methods for Real-World Applications}

\begin{document}


\maketitle

\begin{abstract}
In this paper, we point out a fundamental property of the objective in reinforcement learning, with which we can reformulate the policy gradient objective into a perceptron-like loss function, removing the need to distinguish between on and off policy training. 
Namely, we posit that it is sufficient to only update a policy $\pi$ for cases that satisfy the condition $A(\frac{\pi}{\mu}-1)\leq0$, where $A$ is the advantage, and $\mu$ is another policy. 
Furthermore, we show via theoretic derivation that a perceptron-like loss function matches the clipped surrogate objective for PPO. With our new formulation, the policies $\pi$ and $\mu$ can be arbitrarily apart in theory, effectively enabling off-policy training. To examine our derivations, we can combine the on-policy PPO clipped surrogate (which we show to be equivalent with one instance of the new reformation) with the off-policy IMPALA method. We first verify the combined method on the OpenAI Gym pendulum toy problem. Next, we use our method to train a quadrotor position controller in a simulator. Our trained policy is efficient and lightweight enough to perform in a low cost micro-controller at a minimum update rate of 500 Hz. For the quadrotor, we show two experiments to verify our method and demonstrate performance: 1) hovering at a fixed position, and 2) tracking along a specific trajectory. In preliminary trials, we are also able to apply the method to a real-world quadrotor.
\end{abstract}

\section{Introduction}
\label{Introduction}
Reinforcement learning (RL), and more recently deep reinforcement learning (DRL), has helped put machine learning in the spotlight in recent years \cite{silver2016mastering, silver2017mastering, OpenAI_dota}. Of the many approaches to solve RL and DRL problems, policy gradient methods are a family of model-free algorithms that are widely used and well established \cite{sutton2000policy,mnih2016asynchronous}. Methods that fall into this categorization basically seek to maximize some performance objective with respect to the parameters of a parameterized policy, typically through gradient ascent.

Despite the rigorous theory behind policy gradients in RL, there are a number of practical issues involved with policy gradients in DRL \cite{henderson2018did,ilyas2018deep}. We will focus our analysis on two major DRL improvements for policy gradients in the scope of this paper, the first being Trust Region Policy Optimization (TRPO) \cite{schulman2015trust}, and the second being Proximal Policy Optimization (PPO) \cite{schulman2017proximal}.

TRPO builds upon previous work by first establishing that through optimizing a surrogate function, the parameterized policy is also guaranteed to improve. Next, a \emph{trust region} is used to confine updates so that the step sizes can be large enough for rapid convergence, but small enough so that each update is guaranteed to monotonically improve on the current policy. Building on the success of TRPO, PPO aims to improve data sampling efficiency
while also simplifying the computationally intensive TRPO with a clipped surrogate function. Both TRPO and PPO are discussed in more detail in subsection \ref{Background}.

While TRPO and PPO are used widely and successfully \cite{houthooft2016vime,ho2016generative,tang2016exploration,imav2018:s_morrison_et_al,tan2018sim}, a recent paper by Ilyas et al. \cite{ilyas2018deep} attempts to investigate how well these DRL methods match the theories for fundamental policy gradient methods. Namely, it is not entirely clear why in modern problems with high-dimensional (or even continuous) state-action spaces, a relatively small sample of trajectories can often be sufficient in training strong policies, especially when policy gradient derivations are based on finite Markov decision processes. We summarize Ilyas et al.'s findings in section \ref{Theory}.

Inspired by Ilyas et al.'s work, we devised a novel update method, resulting in a more simplified perspective on PPO. This simplification allows us to treat PPO as an off-policy method, so that it can be combined with IMPALA \cite{espeholt2018impala}, another off-policy method. We first verify our combined method with the simple OpenAI Gym pendulum problem \cite{brockman2016openai}, which we describe in detail in subsection \ref{Pendulum}. Next, we use our method to train a quadrotor controller in a simulator. Since our method is a simplification of PPO, the trained policy is efficient and lightweight enough to perform in a low cost micro-controller at a minimum update rate of 500Hz. We were able to train policies for two tasks, hovering at a fixed position, and tracking along a specified trajectory. Quadrotor experiment results will be discussed in section \ref{QuadrotorExperiments}. Preliminary test show that the resulting model, with a few modifications, can be applied to a real-world quadrotor.

\section{Notation and Background}
\label{Notation&Background}
In this section, we list the notation for the terms we will use in this paper in subsection \ref{Notation}. Our notation is based mostly on two sources \cite{sutton2000policy,schulman2015trust}. We then briefly review the relevant background knowledge for our proposed method in subsection \ref{Background}.

\subsection{Notation}
\label{Notation}
The standard reinforcement learning framework consists of a learning agent interacting with a Markov decision process (MDP), which can be defined by the tuple $(\mathcal{S}, \mathcal{A}, P^a_{ss'}, R^a_s, \rho_0, \gamma)$, where $\mathcal{S}$ is the state space, $\mathcal{A}$ is the action space, $P^a_{ss'}=Pr(s_{t+1}=s'|s_t=s,a_t=a)$ is the transition probability density of the environment, $R^a_s=Pr(r_{t+1}|s_t=s,a_t=a) \forall{s,s'}\in\mathcal{S},a\in\mathcal{A}$ is the set of expected rewards, $\rho_0:\mathcal{S}\rightarrow\mathbb{R^+}$ is the distribution of the initial state $s_0$, and $\gamma\in(0,1)$ is the discount factor. The agent chooses an action at each time step according to the policy $\pi(a|s;\theta)=Pr(a|s;\theta)$. In the modern deep reinforcement learning approach, the policy is represented by a neural network and its weights $\theta$, where $\theta\in\mathbb{R}^{N_\theta}$. 
We will refer to the policy simply as $\pi(a|s)$ for the remainder of this paper.

Let $\eta$ be a function that measures the expected discounted reward with respect to $\pi$, formulated as
\begin{equation}
\label{equ:etaDefinition}
    \eta(\pi)=E\left[\sum_{t\geq1}\gamma^{t-1}r_t \middle| s_0\sim\rho_0,\pi\right].
\end{equation}

We also use the standard definition of the state value function $V^\pi(s)$, state-action value function $Q^\pi(s,a)$, and the advantage function $A^\pi(s,a)$, written as follows:
\begin{equation}
\label{equ:valueDefinition}
    V^\pi(s)=E\left[\sum_{l\geq0}\gamma^l r_{t+l} \middle| s_t=s,\pi \right]
\end{equation}
\begin{equation}
\label{equ:qDefinition}
    Q^\pi(s,a)=E[r_t+\gamma V^\pi(s_{t+1})|s_t=s,a_t=a,\pi]
\end{equation}
\begin{equation}
\label{equ:advantageDefinition}
    A^\pi(s,a)=Q^\pi(s,a)-V^\pi(s)
\end{equation}

The following is the expected return for an agent following the policy $\pi$, expressed in terms of the advantage for another policy $\mu$ \cite{schulman2015trust}:
\begin{equation}
\label{etaForPi&Mu}
    \eta(\pi)=\eta(\mu)+\sum_s \rho^\pi(s) \sum_a \pi(a|s)A^\mu(s,a),
\end{equation}

where $\rho^\pi(s)=\sum_{t\geq0} \gamma^t Pr(s_t=s|\pi)$ is the discounted weighting of states encountered starting at $s_0\sim\rho_0(s)$, then following the policy $\pi$ at each time step.

\subsection{Background}
\label{Background}
The policy gradient idea was first proposed by Sutton et al. \cite{sutton2000policy} as an approach where the policy can be expressed as a function approximator $\pi(a|s)$, which is different from previous value-based RL theory in that it is independent from the various forms of value functions. Consequently, the policy can be improved by updating strictly according to the gradient of the expected reward with respect to the policy's parameters as follows:
\begin{equation}
    \theta \leftarrow \theta+\alpha\frac{\partial\eta(\pi)}{\partial\theta}.
\end{equation}

The so-called policy gradient theorem, derived from Equation \ref{equ:etaDefinition}, states that the gradient of the expected reward can be estimated using a state-action-value (Equation \ref{policyGradientTheoremForQ}) or advantage function (Equation \ref{policyGradientTheoremForA}) approximator:

\begin{equation}
\label{policyGradientTheoremForQ}
    \frac{\partial\eta(\pi)}{\partial\theta} =\sum_{s\in\mathcal{S}}\rho^\pi(s) \sum_{a\in\mathcal{A}}\frac{\partial\pi(a|s)}{\partial\theta}Q^\pi(s,a)
\end{equation}

\begin{align}
\label{policyGradientTheoremForA}
    \frac{\partial\eta(\pi)}{\partial\theta} &=\sum_{s\in\mathcal{S}}\rho^\pi(s) \sum_{a\in\mathcal{A}}\frac{\partial\pi(a|s)}{\partial\theta}(Q^\pi(s,a)-V^\pi(s,a)) \nonumber \\
    &=\sum_{s\in\mathcal{S}}\rho^\pi(s) \sum_{a\in\mathcal{A}}\frac{\partial\pi(a|s)}{\partial\theta}(A^\pi(s,a))
\end{align}


Since the proof for the theorem is covered in detail in Sutton's original policy gradient paper \cite{sutton2000policy}, we will not repeat it here. We refer to these two formulations as on-policy policy gradient in this paper, because the gradient direction $\frac{\partial\pi}{\partial\theta}$ is weighted by the two terms $\rho^\pi(s)$ and $Q^\pi(s,a)$, where the current policy $\pi$ is necessary in order to arrive at an acceptable estimation for $\rho^\pi$ through sampling.

The first off-policy policy gradient method was proposed by Degris et.al. \cite{degris2012off}. In the context of off-policy learning, the original policy $\pi$ is referred to as a target policy, whereas another policy $\mu$, referred to as the behavior policy, is used to collect data for training. Ideally, we want to optimize according to Equation \ref{etaForPi&Mu}, however, $\rho^\pi(s)$ cannot be easily evaluated in an off-policy setting. Instead, we consider an approximate gradient defined by
\begin{equation}
\hat{g}=\sum_{s\in\mathcal{S}}\rho^\mu(s)\sum_{a\in\mathcal{A}}\frac{\partial\pi(a|s)}{\partial\theta}Q^\pi(s,a).
\end{equation}

Degris et al. \cite{degris2012off} state that the policy improvement theorem guarantees that updates along this approximate gradient's direction is not descending. However, regardless of whether an off-policy or an on-policy method is used, it can be quite difficult to apply classical optimization techniques, since any estimation of the true value of $\eta(\pi)$ can be extremely expensive. For example, when performing, say, line search \cite{NoceWrig06} for determining the update step size, $\eta(\pi)$ needs to be evaluated many times.

To address this problem, Schulman et al. \cite{schulman2015trust} investigated using the trust region method to obtain the largest update step size while guaranteeing gradient ascent. The basic idea of the resulting TRPO method is to define a linear approximation of the main objective:
\begin{equation}
    L_\mu(\pi)=\eta(\mu)+\sum_{s\in\mathcal{S}}\rho^\mu(s)\sum_{a\in\mathcal{A}}\pi(a|s)A^\mu(s,a).
\end{equation}
Theoretically, the estimation error can be bounded by 
\begin{gather}
    \beta D^{max}_{KL}(\mu, \pi) \text{, where} \nonumber \\
    \beta=\frac{4max|A^\mu|\gamma}{(1-\gamma)^2}.
\end{gather} 

We can then say that if
\begin{equation}
\label{theoreticalTRPO}
    \sum_{s\in\mathcal{S}}\rho^\mu(s)\sum_{a\in\mathcal{A}}\pi(a|s)A^\mu(s,a)-\beta D^{max}_{KL}(\mu , \pi)\geq 0,
\end{equation}
then
\begin{equation}
    \eta(\pi)\geq\eta(\mu).
\end{equation}
That is, the expected reward for the target policy is higher than the behavior policy. Therefore, by maximizing the left-hand term in Equation \ref{theoreticalTRPO}, we are guaranteed to improve the true objective. Note that in the original derivations by Schulman et al. \cite{schulman2015trust}, $\mu$ and $\pi$ need to be sufficiently similar for this guarantee to hold. From this perspective, while we do refer to policies $\pi$ and $\mu$, TRPO is in practice an on-policy method, where $\pi$ can be thought of as an incremental improvement over $\mu$ after one policy gradient update.

However, it is difficult to directly optimize Equation \ref{theoreticalTRPO}, so the following heuristic approximation is used instead:
\begin{equation}
\begin{split}
    \text{maximize} \quad &L_\mu(\pi)\\
    s.t. \quad &\bar{D}^\rho_{KL}(\mu , \pi)\leq \epsilon
\end{split}
\end{equation}

where $\bar{D}^\rho_{KL}(\mu , \pi) = E_{s\sim\mu}\left[D_{KL}(\mu\parallel\pi)\right]$.

The next algorithm, PPO, tries to keep the first order approximation for $\eta(\pi)$ but proposes a new objective:
\begin{multline}
    \label{PPOclipping}
    L^{CLIP}(\pi)=\\
    min\left[\frac{\pi(a|s)}{\mu(a|s)}\hat{A}, clip\left(\frac{\pi(a|s)}{\mu(a|s)}, 1-\epsilon, 1+\epsilon\right)
    \hat{A}\right]
\end{multline}
where $\hat{A}$ is the estimation of $A^\mu$; e.g. generalized advantage estimation (GAE) \cite{schulman2015high} is often used in practice. By replacing the TRPO surrogate function (Equation \ref{theoreticalTRPO}) with the clipped function (Equation \ref{PPOclipping}), we are essentially excluding samples where the term $\frac{\pi(a|s)}{\mu(a|s)}\geq{1+\epsilon}$ for positive values of $\hat{A}$, or $\frac{\pi(a|s)}{\mu(a|s)}\leq{1-\epsilon}$ for negative values of $\hat{A}$. This is a lower bound to the actual expected reward, so intuitively, by optimizing Equation \ref{PPOclipping}, we are in turn improving the policy. At the same time, Equation \ref{PPOclipping} is less expensive to optimize as the TRPO surrogate function.

\section{Reformulation of the Policy Gradient Objective}
\label{Theory}
Recently, Ilyas et al. \cite{ilyas2018deep} performed a series of experiments and mathematical analyses to investigate how well current deep policy gradient methods, particularly PPO, adhere to the theoretical basis for policy gradients. What they found, in short, were the following:
\begin{itemize}
    \item The gradient estimates for deep policy gradient methods tend to be poorly correlated with the true gradients, and also between each update. Also, as the training progresses or as the task becomes more complex, the gradient estimate also decreases in quality.
    \item Value networks also tend to produce inaccurate predictions for the true value function. They confirm that by using value predictions as a baseline, the variance of the advantage (and in turn the gradient) does decrease. However, compared with no baseline, the improvement is marginal.
    \item The optimization landscape that is produced by deep policy gradient methods often do not match the true reward landscape.
    \item The authors list the above findings as potential reasons why trust region methods require sufficiently similar policies to work. However, since these above factors are not formally considered in the theory for trust regions, it can be difficult to understand why deep policy gradients work, or attribute causes when they do not work.
\end{itemize}
While these findings seem to be critical towards deep policy gradients, the fact is that empirically, these methods are effective in training working policies. To understand better why policy gradients work in the current setting, we wish to take a closer look.

\subsection{Revisiting TRPO and PPO}
\label{TRPO&PPO}
First, to analyze the ideal update scheme for TRPO, we assume the function is arbitrarily perturbed after each batch of training data, say $\pi(a|s)=\mu(a|s)+\delta(a,s)$. Since $\pi$ and $\mu$ are normalized probability density functions, the following condition holds automatically

$$
\sum_{a\in\mathcal{A}} \delta(a,s)=0.
$$

By substituting this definition of $\pi$ into Equation \ref{theoreticalTRPO}, we obtain
\begin{equation*}
\sum_{s\in\mathcal{S}}\rho^\mu(s)\sum_{a\in\mathcal{A}}\delta(a,s)A^\mu(s,a)-\beta D^{max}_{KL}(\mu , \mu+\delta),
\end{equation*}
since $\sum_{a\in\mathcal{A}}\mu(a|s)A^\mu(s,a)=0$. In order to simplify calculation, we consider the lower bound of this formula as follows.
\begin{align*}
    \sum_{s\in\mathcal{S}} \left[\left(\rho^\mu(s)\sum_{a\in\mathcal{A}}\delta(a,s)A^\mu(s,a)\right) - \beta D_{KL}(\mu \parallel \mu+\delta) \right]
\end{align*}
For a small $\delta$ on a given $s$, we expand $D_{KL}(\mu\parallel\mu+\delta)$ around $\delta=0$, which leads to
\begin{align*}
    D_{KL}(\mu\parallel\mu+\delta)&=\sum_{a\in\mathcal{A}}\mu \left[\ln\mu-\ln(\mu+\delta) \right]\\
    &=\sum_{a\in\mathcal{A}}\mu\left[\frac{-\delta}{\mu}+\frac{\delta^2}{2\mu^2}+O(\delta^3)\right]\\
    &\approx \sum_{a\in\mathcal{A}}\frac{\delta^2}{2\mu}.
\end{align*}
Hence, we obtain
\begin{multline*}
    \sum_{s\in\mathcal{S}}\left[\left(\rho^\mu(s)\sum_{a\in\mathcal{A}}\delta(a,s) A^\mu(s,a)\right)-\beta D_{KL}(\mu\parallel\mu+\delta)\right]\\ 
    \approx \sum_{s\in\mathcal{S}, a\in\mathcal{A}}\left[ \rho^\mu(s)A^\mu(s,a)\delta(a,s)-\beta\frac{\delta^2(a,s)}{2\mu}\right],
\end{multline*}
where the first term is linear and the second is quadratic with respect to $\delta$. This relation shows that non-decreasing updates generally exist at the corresponding optimal update, 

\begin{equation}
\label{equ:deltaClosedForm}
    \delta(a,s) = \rho^\mu(s)A^\mu(s,a)\frac{\mu(a|s)}{\beta}.
\end{equation}
From the perspective of policy gradients, a single transition policy update at a given state $s$ and action $a$ is 
\begin{align}
\label{equ:deltaPolicyGradient}
    \delta(a,s) &= \pi(a|s)-\mu(a|s) \nonumber \\
    &= A^\mu(s,a)\frac{\partial\mu(a|s)}{\partial\theta}\Delta\theta+O(\Delta\theta^2) \nonumber  \\
    &\approx A^\mu(s,a)\frac{\partial\mu(a|s)}{\partial\theta}\Delta\theta,
\end{align}
where $\Delta\theta = \alpha \frac{\partial\mu}{\partial\theta}$ and $\alpha$ is the learning rate. Both Equations \ref{equ:deltaClosedForm} and \ref{equ:deltaPolicyGradient} indicate that the update, disregarding the amount, is $\frac{A^\mu}{|A^\mu|}$; i.e. 1 for positive $A^\mu$ and -1 for negative $A^\mu$. In other words, during updates, we encourage actions with positive advantages, and discourage actions for negative advantage. 

The estimation for $|\rho^\mu(s)A^\mu(s,a)|$ when following the function approximation approach is extremely important. This term effectively balances the effects of all the encouraging/discouraging for every state-action pair within a mini-batch, and overall provides limitations to the function approximator. However, as mentioned earlier in this section, the findings by Ilyas et al. \cite{ilyas2018deep} show that modern policy gradient methods can effectively handle DRL problems despite not being able to truly evaluate the value for $|\rho^\mu(s)A^\mu(s,a)|$. To shed light on why it is not necessary to have a highly accurate evaluation of this term, we will need to examine the relation between $\eta(\pi)$ and $\eta(\mu)$.

\subsection{Perceptron-like Objective}
\label{KC-derivation}
We first modify Equation \ref{etaForPi&Mu} by swapping $\pi$ and $\mu$
\begin{equation}
\label{equ:swappedEtaForPi&Mu}
    \eta(\mu) = \eta(\pi) + \sum_{s\in\mathcal{S}}\rho^\mu(s)\sum_{a\in\mathcal{A}}\mu(a|s)A^\pi(s,a).
\end{equation}
By performing algebraic calculations on Equation \ref{etaForPi&Mu} and \ref{equ:swappedEtaForPi&Mu} respectively, we obtain
\begin{align}
\label{equ:etaDifference}
    \eta(\pi)-\eta(\mu)=\sum_{s\in\mathcal{S}}\rho^\pi(s)\sum_{a\in\mathcal{A}}\pi(a|s)A^\mu(s,a)\\
\label{equ:etaDifferenceSwapped}
    \eta(\pi)-\eta(\mu)= -\sum_{s\in\mathcal{S}}\rho^\mu(s)\sum_{a\in\mathcal{A}}\mu(a|s)A^\pi(s,a)
\end{align}
We then add $$-\sum_{s\in\mathcal{S}}\rho^\pi(s)\sum_{a\in\mathcal{A}}\mu(a|s)A^\mu(s,a)$$ to the right hand side of Equation \ref{equ:etaDifference}, and $$\sum_{s\in\mathcal{S}}\rho^\mu(s)\sum_{a\in\mathcal{A}}\pi(a|s)A^\pi(s,a)$$ to Equation \ref{equ:etaDifferenceSwapped}. These two external terms are zero (since $A^\pi$ for the policy $\pi$ is zero), therefore we do not need to balance the equations. Thus, we obtain the following. 
\begin{align}
    \eta(\pi)-\eta(\mu)=\sum_{s\in\mathcal{S}}\rho^\pi(s)\sum_{a\in\mathcal{A}}\left[\pi(a|s)-\mu(a|s)\right]A^\mu(s,a)\\
    \label{equ:newEtaDifference}
    \eta(\pi)-\eta(\mu)= \sum_{s\in\mathcal{S}}\rho^\mu(s)\sum_{a\in\mathcal{A}}\left[\pi(a|s)-\mu(a|s)\right]A^\pi(s,a)
\end{align}
The values for $\rho^\cdot(s)$ are non-negative, therefore these two relations indicate that if one of following inequalities holds for all $(s,a)$ then $\eta(\pi) \geq \eta(\mu)$ (i.e. the expected reward for $\pi$ dominates that of $\mu$):
\begin{enumerate}
    \item $(\pi(a|s)-\mu(a|s))A^\mu(s,a)\geq0$
    \item $(\frac{\pi(a|s)}{\mu(a|s)}-1)A^\mu(s,a)\geq0$
    \item $(\pi(a|s)-\mu(a|s))A^\pi(s,a)\geq0$
    \item $(\frac{\pi(a|s)}{\mu(a|s)}-1)A^\pi(s,a)\geq0$
\end{enumerate}
We refer to these four conditions as dominating inequalities. Note that, first, by following these conditions, we are not optimizing the policies, but rather only guaranteeing in this case that $\pi$ is better than $\mu$. Second, the implication from this result is that we do not require the estimation of the advantage ($A^\pi$ or $A^\mu$) to be highly accurate; we only need to have the correct sign. Finally, when we say this derivation depends on having observed all $(s,a)$, this requirement is on par with the original policy gradient theory. To enforce any one of the above inequalities, we can use a perceptron-like approach \cite{rosenblatt1958perceptron}, e.g.
\begin{equation}
\label{equ:perceptronLike}
    maximize_\theta ~~ min \left[\left(\frac{\pi(a|s)}{\mu(a|s)}-1\right)A^\mu(s,a), \xi\right],
\end{equation}
where $\xi$ is a user-defined margin. It is apparent that this form is equivalent to the PPO clipping surrogate function. We show this from observation. We can enumerate three different cases:

\begin{enumerate}
    \item $A^\mu>0$ and $L^{CLIP}(\pi)\neq\frac{\pi(a|s)}{\mu(a|s)}A^\mu$. This implies that
    \begin{equation*}
        \frac{\pi(a|s)}{\mu(a|s)}A^\mu > (1+\epsilon)A^\mu
    \end{equation*}
    This inequality is equivalent to the case
    \begin{equation}
    \label{equ:clippedSurrogateEquivalent}
        \left(\frac{\pi(a|s)}{\mu(a|s)}-1\right)A^\mu > \epsilon|A^\mu|
    \end{equation}
    
    \item $A^\mu<0$ and $L^{CLIP}(\pi)\neq\frac{\pi(a|s)}{\mu(a|s)}A^\mu$. This implies that
    \begin{equation*}
        \frac{\pi(a|s)}{\mu(a|s)}A^\mu > (1-\epsilon)A^\mu.
    \end{equation*}
    This inequality is equivalent to case 1 (Equation \ref{equ:clippedSurrogateEquivalent}).

    \item $L^{CLIP}(\pi)=\frac{\pi(a|s)}{\mu(a|s)}A^\mu$. The update will be
    \begin{equation*}
        \theta \leftarrow \theta + \frac{\alpha A^\mu}{\mu(a|s)}\frac{\partial \pi(a|s)}{\partial\theta}
    \end{equation*}
\end{enumerate}

In short, it is possible to encourage/discourage actions without using data distribution and trust region assumptions, where the rule conforms to a perceptron-like objective with $\xi=\epsilon|A^\mu|$.
It is also possible to gain an intuition for this new objective: for a specific action $a_t$ at state $s_t$, $\pi$ will be adjusted according to the experience gained through following the policy $\mu$. More specifically, a positive $A^\mu{(s_t,a_t)}$ implies that $a_t$ is believed to be a good action for $\mu$; for case 3, $\pi$ is not as "smarter" as $\mu$, and so it should be adjusted so that $a_t$ is encouraged. For the clipped cases (e.g. case 1), $\pi$ is already better in that it will choose a good action more frequently than $\mu$, so no adjustment will be made.

On the other hand, with function approximators, it can be exceedingly difficult to satisfy the large number of conditions listed above. It is equivalent to asking a function with parameter $\theta\in\mathbb{R}^{N_\theta}$ to satisfy $|\mathcal{S}\times\mathcal{A}|$ number of conditions. However, this can be thought of as a limitation that depends on the capacity of the neural network, and our conjecture is that DRL problems can be solved by only considering one of the four conditions listed above.

Lastly, by reformulating Equation \ref{equ:newEtaDifference}, we can recover the on/off-policy policy gradient formula. 
\begin{multline}
    \sum_{s\in\mathcal{S}}\rho^\mu(s)\sum_{a\in\mathcal{A}}\left[\pi(a|s)-\mu(a|s)\right]A^\pi(s,a)\\
    =\sum_{s\in\mathcal{S}}\rho^\mu(s)\sum_{a\in\mathcal{A}}\left[\pi(a|s)-\mu(a|s)\right]Q^\pi(s,a), 
\end{multline}
since $V^\pi(s)\sum_a\left[\pi(a|s)-\mu(a|s)\right]=0$. We then take the derivative with respect to $\theta$ to obtain
\begin{multline}
    \frac{\partial \eta(\pi)}{\partial \theta} = \sum_{s\in\mathcal{S}} \rho^\mu(s)\sum_{a\in\mathcal{A}}\frac{\partial \pi(a|s)}{\partial\theta}Q^\pi(s,a)\\
    +\sum_{s\in\mathcal{S}} \rho^\mu(s)\sum_{a\in\mathcal{A}}\left[\pi(a|s)-\mu(a|s)\right]\frac{\partial Q^\pi(s,a)}{\partial\theta}.
\end{multline}
The first term is exactly the same as the off-policy policy gradient proposed by Degris et.al. \cite{degris2012off}. We conclude that the off-policy policy gradient is indeed an approximation, where the accuracy is determined by the difference between $\pi$ and $\mu$. The act of recovering the on-policy policy gradient is to simply send $\mu$ to approach $\pi$.

\section{Algorithm}
\label{Algorithm}

\begin{algorithm}[tb]
  \caption{Learner}
  \label{alg:LearnerThread}
\begin{algorithmic}
  \State wait learner thread lock
  \Repeat
  \State sample $T \in \mathbf{B}$
  \State $tmpA=0$
  \For{$i=n-1$ {\bfseries to} $0$}
  \State $A_i^{trace} = \left(r_i + \gamma V_{i+1} - V_i\right) + \gamma*tmpA$
  \State $tmpA=\min(1, \frac{\pi_i}{\mu_i})A_i^{trace}$
  \State $V_i^{trace} = V_i + tmpA$
  \EndFor
  \State $\theta_a \leftarrow \theta_a + \alpha \frac{\partial L_{policy} }{\partial\theta}$
  \State $\theta_v \leftarrow \theta_v - \alpha \frac{\partial L_{value}}{\partial\theta}$
  \Until{Thread terminate}
\end{algorithmic}
\end{algorithm}

\begin{algorithm}[tb]
   \caption{Runner}
   \label{alg:RunnerThread}
\begin{algorithmic}
   \State {\bfseries Input:} $\mathbf{B}$ replay buffer, Learner thread, $\tilde{M}$ maximum trajectories, $n$ trajectories length
   \State Initialize $B$, Learner thread lock, $\theta_a$ policy weight, $\theta_v$ value function weight
   \For{$j=0$ {\bfseries to} $\tilde{M}-1$}
   \If{$j > 10$} 
   \State release learner thread lock
   \EndIf
   \State sample $T_j$=$(s_i, a_i, \pi(a_i|s_i), r_i, s_{i+1})_j \sim \pi$
   \State $\mathbf{B}$ = $\mathbf{B} \cup T_j$
   \EndFor
   \State Kill learner thread
\end{algorithmic}
\end{algorithm}

With its clipped surrogate function, PPO is effectively a "nearly on-policy" algorithm in that it does not follow a policy's sample distribution exactly. With the reformulation of the RL objective, we can try to use PPO as an off-policy method. To do this, we use a replay buffer. In an off-policy setting, we are not able to access the estimation $V^\mu$. Therefore, we choose to use the fourth dominating condition
\begin{equation*}
\left(\frac{\pi(a|s)}{\mu(a|s)}-1\right)A^\pi\geq \epsilon|A^\pi|.
\end{equation*}

However, the estimation of $A^\pi$ is still difficult, so we turn to using V-trace, a state-value estimation technique used in IMPALA \cite{espeholt2018impala}, for value function training, where all parameters, e.g., $\bar{c}, \bar{\rho}$, are set to be $1$. We define $A^{trace}$ to be
\begin{equation}
    A^{trace}_t = A_t + \gamma\min (1,\frac{\pi_{t+1}}{\mu_{t+1}})A^{trace}_{t+1},
\end{equation}
where $A_t=r_t + \gamma V_{t+1}- V_{t}$, and $V_t$ is the estimation from the current value network, similar to the settings published by Rolnick et al. \cite{rolnick2018experience}. 
Let $V^{trace}$ be defined from $A^{trace}$ accordingly. 
Combined together, the objective is the following
\begin{equation}
    \begin{split}
        L_{policy} &= \sum_{(s,a) \in T} \min\left[\left(\frac{\pi(a|s)}{\mu(a|s)}-1\right)A^{trace}, \epsilon|A^{trace}| \right]\\
        L_{value} &= \frac{1}{|T|} \sum_{(s,a) \in T}\left(V(s)-V^{trace}\right)^2.
    \end{split}
\end{equation}
In all of our experiments (including the quadrotor experiments in the next section), a mini-batch is set to be $200$ time steps. The corresponding pseudo-code is written in Algorithm \ref{alg:LearnerThread}. It is worth mentioning that the parameters are shared in the same memory location during the entire training process. Algorithm \ref{alg:RunnerThread} does the trajectory collection, running concurrently with the learner thread. 

\begin{figure}
    \centering
    \includegraphics[width=9cm, height=4.8cm]{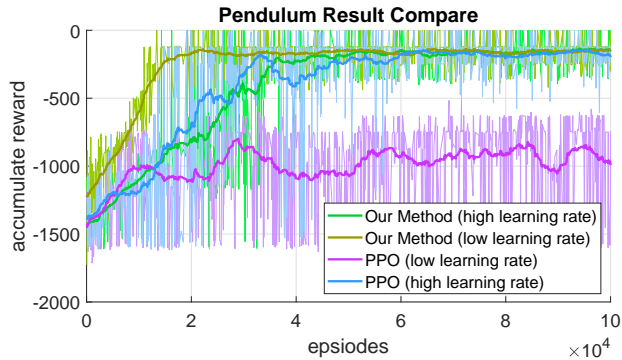}
    \caption{Learning curves for the pendulum problem. The high learning rate is set to be $10^{-4}$, while the lower learning rate is set to be $0.5\times10^{-5}$.}
    \label{fig:pendulum}
\end{figure}

\subsection{Pendulum}
\label{Pendulum}
To verify the algorithm, we first apply our method to the OpenAI pendulum environment. 
As shown in Figure \ref{fig:pendulum}, compared to a minimally tuned PPO, our combined method is able to achieve slightly better performance, though the reformulated objective is nearly equivalent to PPO with slightly relaxed criteria. The pendulum task is time-unlimited, but in the experiment each episode is a finite number of steps. To address this potential terminal state definition issue, we perform partial episode bootstrapping \cite{Pardo2018TimeLI}. We believe the combined on/off-policy method should perform better with proper hyperparameter selection. Since the pendulum experiment is merely a simple test to verify our hypothesis, we leave this as future work.

\section{Quadrotor Experiments}
\label{QuadrotorExperiments}


\begin{figure}
    \centering
    \includegraphics[width=8cm, height=4.8cm]{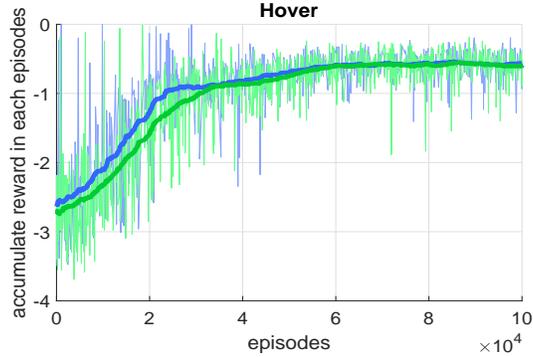}
    \caption{Learning curve for optimizing the hovering policy in two training trials.}
    \label{fig:hover_reward}
\end{figure}

Next, we apply the combined on/off-policy method to train a model-free quadrotor control agent. For the scope of this paper, we simply discuss training and testing the DRL agent in a simulator, though preliminary experiments show that the learned model can be applied to a real-world scenario.

To train and verify our algorithm in a simulated environment, we construct a quadrotor simulator written in Python based on the dynamic model outlined by Bangura et al. \cite{bangura2012nonlinear}. We use the simplest dynamic model, which only calculates the effects of gravity and the forces generated by the motors in our simulator.

\subsection{Experiment Setting}
\label{Exp_setting}

In this experiment, we focus on learning two tasks: 1) position control, and 2) tracking a specific trajectory. For the first task, the DRL agent needs to learn to stabilize the quadrotor and hover at a defined position. For the second task, the DRL agent needs to fly along a circular trajectory with a constant speed.

During training, the target location is set to be the origin of the space, where the goal for the quadrotor agent is to reach and stabilize at the origin. The quadrotor initial state (position, velocity, angle, and angular velocity) is randomized.

To increase flexibility of the controller, we add an offset force $F_o$ on each motor during the training process as described in Equation \ref{forceOutput}. The sum of the offset force and policy network output $F_\pi$ are used as the control output on the quadrotor.
\begin{equation}
\begin{aligned}
    F&=F_o+F_\pi\\
    F_o&=\frac{1}{4}mg\label{forceOutput}
\end{aligned}
\end{equation}
By using this method, the machine learned action chosen by the policy only needs to adjust the motor thrust around a reference action, rather than exploring a wide action space. Empirically, this method results in a much faster training process than without a reference action.

\begin{figure}
    \centering
    \includegraphics[width=8cm, height=4.8cm]{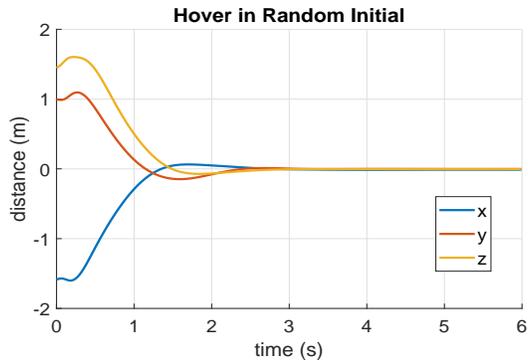}
    \caption{The quadrotor's $x$, $y$, and $z$ distance from the origin $[0,0,0]$ in a hovering trial, starting from a random initial state with position $[-1.58,0.99,1.45]$ and velocity $[-2.39,6.20,0.77]$.}
    \label{fig:hover_result}
\end{figure}

With the aforementioned method, the two tasks were evaluated in our experiments, where each episode consists of at most 200 steps, and each step takes 0.01 second in the simulation environment. The experiments for the two tasks are described in the subsequent subsections respectively. 

\subsection{Hovering Task}
\label{Hovering}
The basic function of the quadrotor controller is to hover at a specific position. The reward function is designed as
\begin{equation*}
\begin{aligned}
    \text{reward}=-\begin{bmatrix}
                w_1 & w_2 & w_3
                \end{bmatrix}^T
                \begin{bmatrix}
            \parallel q_e\parallel & \parallel p_e\parallel & \parallel a\parallel
                \end{bmatrix},
\end{aligned}
\end{equation*}
where $q_e$, $p_e$, and $a$ are angle error, position error, and action output. $w_1$ to $w_3$ are the weights of the error. The reward is normalized to $\pm1$ for faster convergence.  

Based on the algorithm described above, we train our model with 10 collected episodes (at most 200 steps in one episode). Two training runs are collected and the training curves are shown in Figure \ref{fig:hover_reward}. 
Using our algorithm, it takes only 10 million time steps to train a policy such that the quadrotor can steadily hover at the designated position. In contrast, in the work by Hwangbo et al. \cite{hwangbo2017control}, which is the first to use RL on quadrotor control to our knowledge, the quadrotor is still unstable after 100 million steps, and appears to achieve a similar level of stability at 2150 million steps\footnote{Based on their video at \hyperlink{https://youtu.be/zIi4yHYJdJY?t=111}{https://youtu.be/zIi4yHYJdJY?t=111}, and the network training section of their paper; they used one million time steps per iteration, and the best policy was trained with 2150 iterations.}.

In addition, Figure \ref{fig:hover_result} shows the results for a randomly initialized trial.

\begin{figure}
    \centering
    \includegraphics[width=8cm, height=4.8cm]{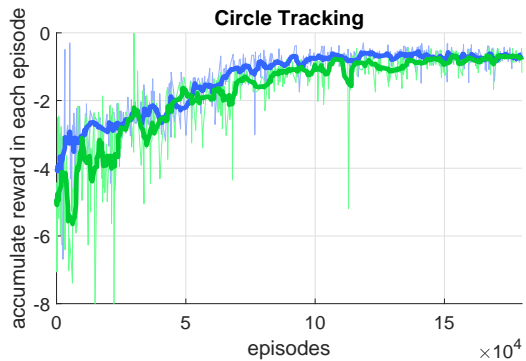}
    \caption{Learning curve for optimizing the action policy in two training trials for the tracking task.}
    \label{fig:circle_reward}
\end{figure}

\subsection{Trajectory Tracking Task}
\label{Tracking}
In addition to hovering at a designated point as a regulation problem, trajectory tracking is another common scenario in quadrotor operation. We create a circular path on the $xy$-plane and train the agent to fly on the circle with constant speed, starting from a random initial condition. The reward function is designed as
\begin{equation*}
\begin{aligned}
    \text{reward}=-\begin{bmatrix}
                w_4 & w_5 & w_6
                \end{bmatrix}^T
                \begin{bmatrix}
                \parallel d\parallel & \parallel \chi \parallel & \parallel a \parallel
                \end{bmatrix}
\end{aligned}
\end{equation*}
where $w_4$ to $w_6$ are the weights of the error, $d$ is the distance to the circular trajectory, and $\chi$ is the direction that the quadrotor is flying on the trajectory. To ensure the quadrotor flies on the desired direction on the path, we take the cross product of the position and velocity on the $xy$-plane, which we call $\chi$, written as:
\begin{equation*}
\begin{aligned}
    \chi=xv_y-yv_x-r_{des}v_{des}
\end{aligned}
\end{equation*}
where $r_{des}$ and $v_{des}$ are the desired radius and quadrotor velocity on the trajectory.

\begin{figure}
    \centering
    \includegraphics[width=8cm, height=5.0cm]{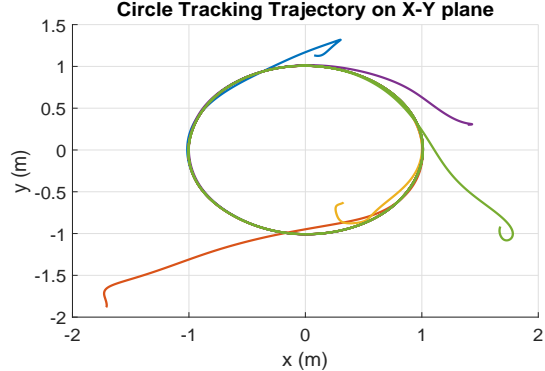}
    \caption{The quadrotor starts from 5 different initial conditions and follows the designed trajectory.}
    \label{fig:circle_pos_result}
\end{figure}

\begin{figure}
    \centering
    \includegraphics[width=8cm, height=4.5cm]{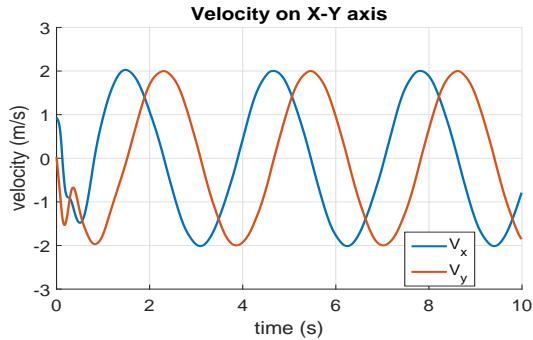}
    \caption{Quadrotor velocity on the axis of $x$ and $y$ with maximum speed 2 m/s.}
    \label{fig:circle_vel_result}
\end{figure}

In our experiment, the radius of the circle is set to be 1 meter and the velocity is 2 m/s. Two training curves are shown in Figure \ref{fig:circle_reward}. The controller learned to fly in a circle after around 1.0 million training episodes. Figure \ref{fig:circle_pos_result} shows five different trials with five different initial states, where the quadrotor is able to fly on our designated 1 m radius circular trajectory. With the constant velocity requirement, a simple harmonic motion on the $x/y$-axes can be observed in Figure \ref{fig:circle_vel_result}. To our knowledge, this is the first work that can train a quadrotor to track along a specific trajectory using RL. 


\section{Conclusion}
\label{Conclusion}

In this paper, we point out a fundamental property of the objective as in subsection \ref{KC-derivation}. Through derivation, we can reformulate the policy gradient objective into a perceptron-like loss function, following what we refer to as the four dominating inequalities. Since the reformulation is based on Equation \ref{etaForPi&Mu}, a general formula with no restrictions on the KL-divergence of $\pi$ and $\mu$, training following this new objective can be both on- or off-policy.

More specifically, we posit that it is sufficient to only update a policy $\pi$ for cases that satisfy the condition $A(\frac{\pi}{\mu}-1)\leq0$. Furthermore, we show via theoretic derivation that a specific instance of our perceptron-like loss function matches the clipped surrogate objective for PPO in each update. 

To examine our derivations, we can combine PPO with IMPALA. The combined method is demonstrated to work well for a quadrotor simulator. Preliminary trials also show that the trained policy can be used on a real world quadrotor. It is somewhat surprising that our trained agent in the above hovering task can be applied in the real world  without any sim-to-real techniques (e.g., randomization or accurate physical models \cite{tan2018sim}). On the simulator, our method was able to hover steadily with 10 million time steps. In previous work by Hwangbo et al. \cite{hwangbo2017control}, the first to use RL on quadrotor control to our knowledge, the quadrotor is still unstable after 100 million steps. Additionally, to our knowledge, this paper is the first work that can train quadrotor to track along a specific trajectory using RL.

Lastly, our derivations could shed light on one of the questions posed by Ilyas et.al \cite{ilyas2018deep}, i.e., why modern policy gradient methods work despite the inaccurate state-value function predictors and policy gradient directions.

\bibliography{main}
\bibliographystyle{unsrt}

\end{document}